\documentclass[letterpaper, 10pt, journal, twoside]{IEEEtran}

\usepackage{amsthm}
\usepackage{times}
\usepackage{multicol}
\usepackage{xcolor}
\usepackage[bookmarks=true,hidelinks]{hyperref}
\usepackage{cite}
\usepackage{amsmath, amssymb}
\usepackage{amsfonts}
\usepackage{graphicx}
\usepackage{siunitx}
\usepackage{standalone}
\usepackage{booktabs}
\usepackage[ruled,vlined,linesnumbered,noend]{algorithm2e}
\usepackage{mdframed}
\usepackage{fancyvrb,multirow}
\usepackage{soul}
\usepackage{dsfont,mathabx}
\usepackage{array, booktabs}
\usepackage{subfigure}
\usepackage{booktabs}
\usepackage{makecell}
\usepackage{threeparttable}
\usepackage{tikz}
\usetikzlibrary{arrows.meta,positioning,fit}
\usetikzlibrary{calc}

\theoremstyle{definition}

\newtheorem{remark}{Remark}

\newcommand{\RALReceivedDate}{June 24, 2026}
\newcommand{\RALRevisedDate}{September 16, 2026}
\newcommand{\RALAcceptedDate}{October 20, 2026}
\newcommand{\RALFinalMonthYear}{October, 2026}

\title{PushAround: Collaborative Path Clearing via Physics-Informed Hybrid Search}

\author{Zili Tang$^{1}$, Zongyuan Li$^{2}$, Zishao Qiao$^{1}$, and Meng Guo$^{1}$%
\thanks{Manuscript received: \RALReceivedDate; Revised: \RALRevisedDate;
Accepted: \RALAcceptedDate.}%
\thanks{This paper was recommended for publication by Editor M. Ani Hsieh
upon evaluation of the Associate Editor and Reviewers' comments.
This work was supported by the National Natural Science Foundation of China
under grants U2241214 and T2121002.
\textit{(Corresponding author: Meng Guo.)}}%
\thanks{The authors are with Peking University, Beijing 100871, China:
$^{1}$School of Advanced Manufacturing and Robotics;
$^{2}$College of Engineering (e-mail: meng.guo@pku.edu.cn).}%
\thanks{Author accepted manuscript.}}

\usepackage{eso-pic}
\newcommand{\IEEEAcceptedNotice}{%
\copyright~2026 IEEE. Personal use of this material is permitted.
Permission from IEEE must be obtained for all other uses, in any current or
future media, including reprinting/republishing this material for advertising
or promotional purposes, creating new collective works, for resale or
redistribution to servers or lists, or reuse of any copyrighted component of
this work in other works.}
\AddToShipoutPictureFG*{%
  \AtPageLowerLeft{%
    \put(\LenToUnit{\dimexpr(\paperwidth-\textwidth)/2\relax},\LenToUnit{9pt}){%
      \parbox[b]{\textwidth}{\fontsize{7}{8}\selectfont\IEEEAcceptedNotice}%
    }%
  }%
}

\hypersetup{
  pdftitle={PushAround: Collaborative Path Clearing via Physics-Informed Hybrid Search},
  pdfauthor={Zili Tang, Zongyuan Li, Zishao Qiao, Meng Guo},
  pdfsubject={Author accepted manuscript, IEEE Robotics and Automation Letters},
  pdfkeywords={Multi-Robot Systems, Task and Motion Planning, Manipulation Planning}
}
\begin{document}
\maketitle

\begin{abstract}
Large vehicles in cluttered environments are often blocked by movable
obstacles, motivating mobile robot teams to clear traversable corridors.
Navigation Among Movable Obstacles (NAMO) approaches often abstract obstacle
displacements while neglecting robot dimensions, obstacle mass, contact
geometry and forces, yielding physically infeasible plans.
\emph{PushAround} is a physics-informed hybrid search that couples geometric
connectivity reasoning with short-horizon simulation to construct executable
clearing plans. Repeated W--clearance connectivity graph (WCCG) queries expose
frontier gaps requiring widening and rank them by estimated downstream costs.
The search generates direct or recursive pushing candidates, selecting
obstacles, directions, contacts and forces, and admits physics-evaluated
candidates.
Numerical experiments validate WCCG against standard connectivity checks and
demonstrate scaling to denser clutter and teams of up to four robots, with
robustness to simulated execution-side physical-parameter mismatch.
\end{abstract}

\begin{IEEEkeywords}
Multi-Robot Systems, Task and Motion Planning, Manipulation Planning.
\end{IEEEkeywords}
\section{Introduction}\label{sec:intro}
\begin{figure}[t!]
  \centering
  \includegraphics[width=0.9\linewidth]{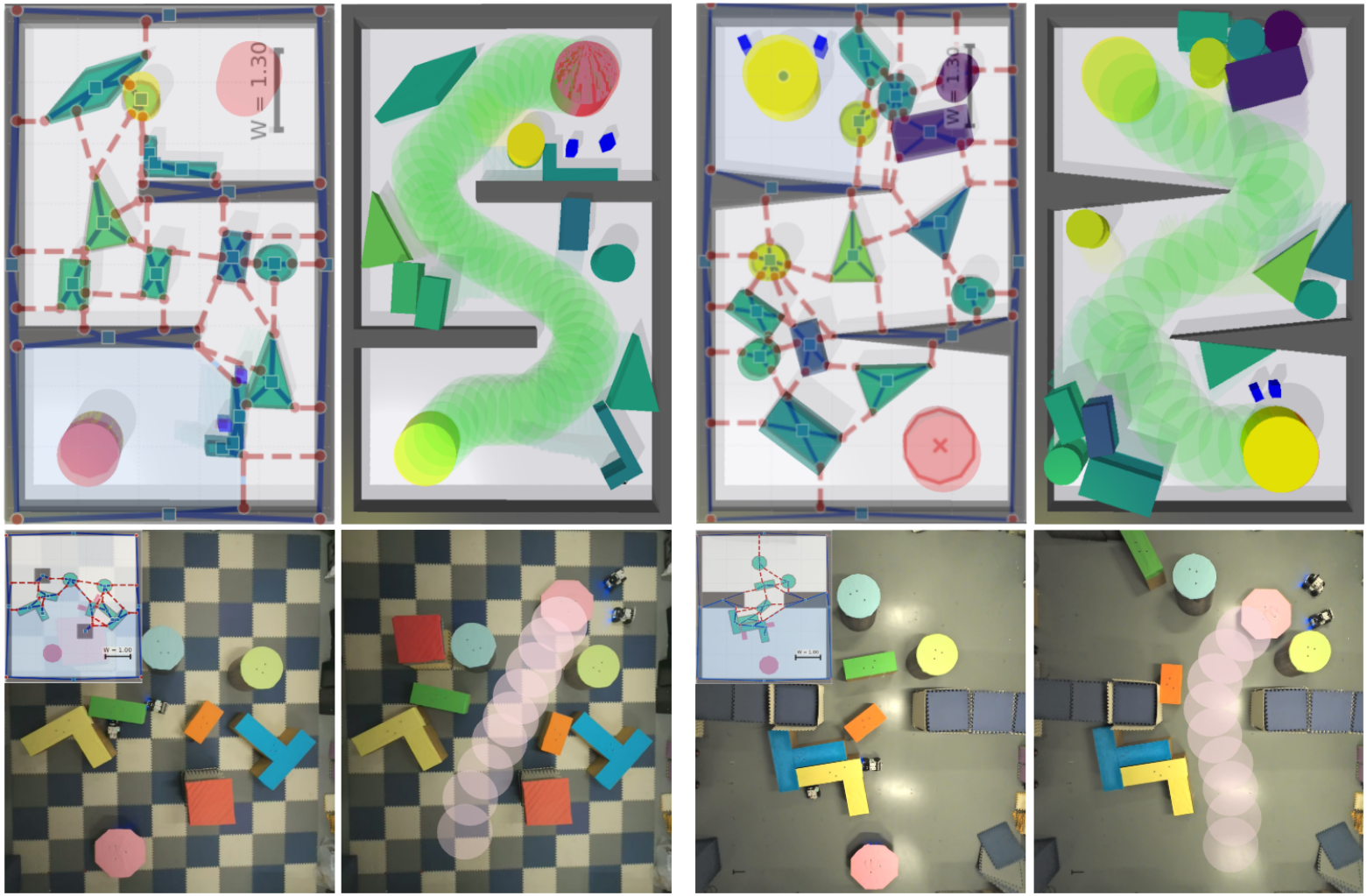}%
  \vspace{-0.1in}
  \caption{
    Collaborative path clearing in simulation (\textbf{Top})
    and hardware (\textbf{Bottom}). Two robots push movable obstacles to
    create a $W$--clear path for a large vehicle from start to goal.
  }
  \vspace{-6mm}
  \label{fig:snapshots}
\end{figure}

\IEEEPARstart{I}{n} warehouses, disaster sites and urban spaces, movable
obstacles such as pallets, boxes and equipment can block large vehicles,
motivating mobile robot teams to clear traversable corridors.
Conventional path planners assume static obstacles and fail once routes are
blocked~\cite{liu2023path}. Navigation Among Movable Obstacles (NAMO) creates
routes by pushing, pulling or rotating obstacles~\cite{stilman2005navigation},
yet many methods abstract away robot dimensions, obstacle size and mass,
contact geometry, forces and dynamics, producing \emph{geometrically valid
but physically unrealizable plans}. As shown in Fig.~\ref{fig:snapshots},
a large robot may be unable to reach obstacle boundaries while a small
robot lacks pushing force. Cooperative pushing by several small robots
offers an alternative, but simultaneous pushes can trigger chained motions
of several obstacles, introducing physical coupling that complicates planning.

\subsection{Related Work}\label{subsec:intro-related}

\subsubsection{Navigation Among Movable Obstacles}
NAMO extends motion planning to cluttered environments by planning obstacle
displacements through pushing, pulling, or
rotating~\cite{stilman2005navigation,stilman2007manipulation}. Later work considers heuristic search, learning-based local planning, physics-based rollouts, and collaborative pushing~\cite{stilman2007manipulation,yao2024local,ren2025search,weeda2025pushing,tang2026pushingbots}.
However, many NAMO formulations simplify robot geometry, actuation limits,
object mass, and contact dynamics, which can yield unrealizable plans.

\begin{figure*}[t!]
  \centering
  \includegraphics[width=0.92\linewidth]{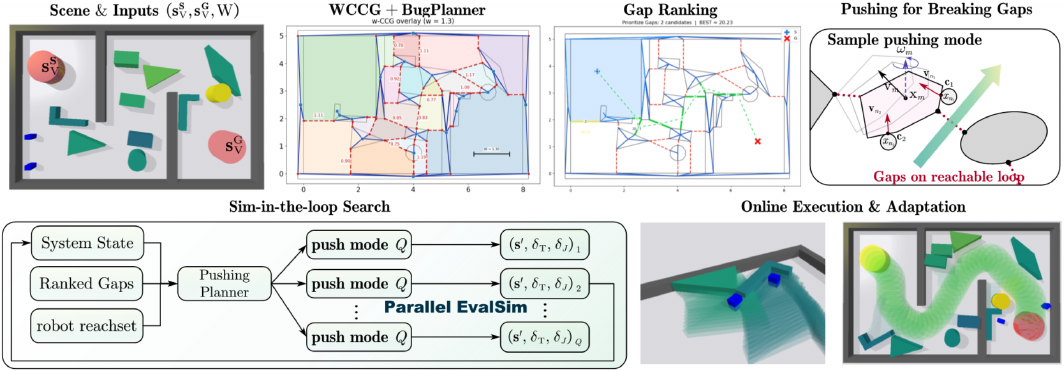}
  \vspace{-0.1in}
\caption{Overview of the proposed framework.
\textbf{Top:} Scene inputs, WCCG construction, gap ranking, and sampling of pushing
modes for breaking frontier gaps;
\textbf{Bottom-left:} Sim-in-the-loop hybrid search with parallel validation in simulation;
\textbf{Bottom-right:} Online execution and adaptation during path clearing.}
  \label{fig:overall}
  \vspace{-5mm}
\end{figure*}

\subsubsection{Collaborative Pushing}
Studies cover pushing mechanics and limit surfaces~\cite{goyal1989limit,lynch1992manipulation}, cooperative transport with robot swarms~\cite{chen2015occlusion}, multi-robot box pushing with reinforcement learning~\cite{wang2006multi}, collaborative pushing of polytopic objects in cluttered scenes~\cite{tang2026pushingbots}, and learning-based loco-manipulation~\cite{feng2025learning}. These studies focus on a prescribed object rather than multi-object corridor clearing.

\subsubsection{Physics-Informed Planning}
Dynamics models, contact reasoning, and simulation feedback inform motion
generation: learned dynamics for humanoid
contact planning~\cite{lin2019efficient}, multi-contact force control~\cite{rouxel2024multi}, demonstration-guided optimal control over contact modes~\cite{xue2023demonstration}, physics simulation interleaved with multi-agent pathfinding~\cite{saxena2023planning,ren2025object}, and learned physical priors~\cite{ni2023progressive,liu2025physics}. These methods typically address a prescribed target or a small set of objects. Corridor clearing instead requires choosing which of many movable obstacles to rearrange so that a larger vehicle can pass.

\subsection{Our Method}\label{subsec:intro-our}

As shown in Fig.~\ref{fig:overall}, this work introduces \textbf{PushAround},
a physics-informed framework for collaborative multi-robot pushing that
constructs traversable corridors for large vehicles in cluttered environments.
Its central novelty is a hybrid search that jointly determines which obstacles
to displace and how to push them, coupling the high-level clearing sequence
with low-level pushing modes specifying contact points, directions and forces.
The framework integrates a W--clearance connectivity graph (WCCG) that certifies
vehicle passage under width constraints, frontier-gap ranking by estimated
effort, and pushing-mode generation under geometric and actuation constraints.
Guided by the selected gap, the search proposes direct pushes or recursively
clears blocking obstacles, and expands the plan through physics-evaluated
transitions that account for robot dimensions, object mass, contact geometry
and coupled pushing dynamics. Execution feedback updates connectivity and
triggers replanning when needed.

The contributions are twofold: {(I)} it introduces a unified framework that
couples collaborative pushing with physics-informed verification,
producing executable plans for path clearing in cluttered environments;
and {(II)} it demonstrates empirical gains in feasibility, efficiency and scalability over evaluated NAMO and
collaborative-clearing baselines, with robustness to execution-side physics
mismatch.

\section{Problem Description}\label{sec:problem}
\subsection{Workspace and Robots}\label{subsec:ws}
The workspace is a bounded planar region $\mathcal{W}\subset\mathbb{R}^2$
containing immovable structures~$\mathcal{O}\subset\mathcal{W}$ and $M\geq0$
movable rigid polygons
$\boldsymbol{\Omega}\triangleq\{\Omega_1,\cdots,\Omega_M\}$.
Each movable obstacle~$\Omega_m\subset\mathcal{W}$ has mass~$\mathsf{M}_m$,
inertia~$\mathsf{I}_m$, frictional parameters and state
$\mathbf{s}_m(t)\triangleq(\mathbf{x}_m(t),\psi_m(t))$, with
$\mathbf{x}_m(t)\in\mathbb{R}^2$ the centroid position and
$\psi_m(t)\in\mathbb{R}$ the orientation; $\Omega_m(t)$ denotes the region it
occupies at time~$t$. A team of~$N$ robots,
$\mathcal{R}\triangleq\{R_1,\cdots,R_N\}$, operates as a cooperative unit. Each
robot~$R_i$ is a rigid body with state
$\mathbf{s}_{R_i}(t)\triangleq(\mathbf{x}_{R_i}(t),\psi_{R_i}(t))$ and
footprint~$R_i(t)$. The instantaneous free space is given by:
\begin{equation}\label{eq:freespace}
\widehat{\mathcal{W}}(t)\triangleq\mathcal{W}\setminus\Big(
\mathcal{O}\cup\bigcup_{m=1}^{M}\Omega_m(t)\cup\bigcup_{i=1}^{N}R_i(t)
\Big).
\end{equation}

\subsection{External Vehicle and Clearance Goal}\label{subsec:vehicle}
An external vehicle~$\texttt{V}$ whose footprint is enclosed by a disk of
radius~$W/2>0$ must navigate within the workspace from a start position
$\mathbf{x}_\texttt{V}^{\texttt{S}}\in\mathcal{W}$ to a goal position
$\mathbf{x}_\texttt{V}^{\texttt{G}}\in\mathcal{W}$. The robots rearrange obstacles to admit this disk. At time~$t$, the
admissible center space is
$\mathcal{F}_W(t)\triangleq
\{p\in\widehat{\mathcal{W}}(t)\mid
\mathbb{B}_{W/2}(p)\subset\widehat{\mathcal{W}}(t)\}$, where
$\mathbb{B}_{W/2}(p)$ is the disk of radius~$W/2$ centered at~$p$. The $W$-clearance condition is:
\begin{equation}\label{eq:wclear}
\exists\,\mathcal{P}^W_\texttt{V}\subset\mathcal{F}_W(t):\
\mathbf{x}_\texttt{V}^{\texttt{S}}\leadsto\mathbf{x}_\texttt{V}^{\texttt{G}},
\end{equation}
which certifies a collision-free centerline path for any vehicle footprint
contained in the enclosing disk.

\subsection{Collaborative Pushing Modes}\label{ss:interaction_mode}
To enlarge blocked gaps, the robots may actively
reconfigure~$\boldsymbol{\Omega}$ by pushing movable obstacles. For obstacle
$\Omega_m\in\boldsymbol{\Omega}$, a pushing mode is denoted by
$\boldsymbol{\xi}_m\triangleq(\mathbf{c}_m,\mathbf{u}_m)$, where
$\mathbf{c}_m=(c_1,\cdots,c_N)$ is an ordered tuple of contact points on
$\partial\Omega_m$, and the ordering specifies the robot-to-contact assignment.
The vector~$\mathbf{u}_m\in\mathbb{R}^{2N}$ stacks the corresponding planar
contact forces, expressed in the body frame of~$\Omega_m$; an inactive
robot carries a null contact $c_i=\varnothing$ with zero force.
For a contact tuple~$\mathbf{c}_m$, the set of admissible forces
${U}_m(\mathbf{c}_m)\subset\mathbb{R}^{2N}$ collects contact forces
satisfying the robot pushing limits, robot--object friction, and unilateral
contact constraints. The body wrench induced by a mode is defined as:
\begin{equation}
\mathbf{q}_m(\boldsymbol{\xi}_m)
\triangleq
G_m(\mathbf{c}_m)\mathbf{u}_m,
\label{eq:contact_wrench_map}
\end{equation}
where~$G_m(\mathbf{c}_m)\in\mathbb{R}^{3\times 2N}$ is the contact-to-wrench mapping;
and $\mathbf{q}_m(\boldsymbol{\xi}_m)\in\mathbb{R}^3$.
The admissible mode set is state dependent as denoted by~$\Xi_m(\mathbf{s})$.
It contains modes whose contacts are reachable from the
current robot configurations, pre-contact configurations are feasible in
the current clutter, and forces satisfy
$\mathbf{u}_m\in{U}_m(\mathbf{c}_m)$. Hence, applying a feasible
mode~$\boldsymbol{\xi}_m\in\Xi_m(\mathbf{s}(t))$ yields the physical
transition:
\begin{equation}\label{eq:transition}
  \mathbf{s}(t^+)\triangleq
  \Phi\big(\mathbf{s}(t),\,m,\,\boldsymbol{\xi}_m\big),
\end{equation}
where~$\mathbf{s}(t)$ stacks all robot and obstacle states.
The force vector~$\mathbf{u}_m$ serves as a feasibility variable to certify
wrench generation under the modeled contact and actuation constraints rather
than a directly commanded control input.

\subsection{Problem Statement}\label{subsec:objective}

The goal is to compute a \emph{hybrid schedule} that reconfigures movable
obstacles to admit a $W$-feasible path for the external vehicle.
The schedule is defined as:
\begin{equation}\label{eq:schedule}
    \pi\triangleq
    \left\{
    \left(
    \Omega_{m_k},
    \mathbf{v}_k,
    \boldsymbol{\xi}_k
    \right)
    \right\}_{k=1}^{K},
\end{equation}
where~$\Omega_{m_k}\in\boldsymbol{\Omega}$ is the manipulated obstacle,
$\mathbf{v}_k$ its pushing direction, $\boldsymbol{\xi}_k\in\Xi_{m_k}$ its
pushing mode. The node cost is:
\begin{equation}\label{eq:problem}
    \mathcal{J}_{\mathrm{clr}}(\pi)
    \triangleq
    T+
    \alpha
    \sum_{k=1}^{K}
    J\left(
    m_k,\boldsymbol{\xi}_k,\mathbf{s}(t_k)
    \right),
\end{equation}
where~$T\triangleq\sum_{k=1}^{K}\Delta t_k$ denotes
the robotic clearance duration excluding vehicle traversal,
with $\Delta t_k>0$ the duration that evaluation in~\eqref{eq:sim} returns for the $k$-th push,
 and~$J(\cdot)$ is the simulation-estimated manipulation cost weighted by $\alpha>0$.
 Vehicle traversal enters only through the terminal $W$-clearance condition in~\eqref{eq:wclear}.
 The constraints in~\eqref{eq:freespace}--\eqref{eq:transition}
 enforce valid pushing transitions.
 The realized clearance cost is defined in~\eqref{eq:problem}; PIHS samples finitely and does not guarantee optimality.

\begin{remark}\label{remark:uniqueness}
  The objective couples obstacle selection, pushing-mode generation and switching in one hybrid planning problem. Its combinatorial-continuous search includes robot-team contacts and forces alongside the obstacle rearrangements considered in NAMO~\cite{stilman2005navigation,stilman2007manipulation}.
  \hfill~$\blacksquare$
\end{remark}

\section{Proposed Solution}\label{sec:solution}

As shown in Figs.~\ref{fig:overall} and~\ref{fig:pihs_overview}, PushAround first
uses the W-Clearance
Connectivity Graph in Sec.~\ref{subsec:wccg} and gap ranking in
Sec.~\ref{subsec:gap} to identify and prioritize blocking gaps. It then
generates and evaluates physically feasible collaborative pushing modes in
Sec.~\ref{subsec:push}. These components are integrated into the
simulation-in-the-loop hybrid search of Sec.~\ref{subsec:simloop} and the online
execution procedure of Sec.~\ref{subsec:overall}.

\subsection{W--Clearance Connectivity Graph}\label{subsec:wccg}

\begin{figure}[t!]
  \centering
  \includegraphics[width=0.95\linewidth]{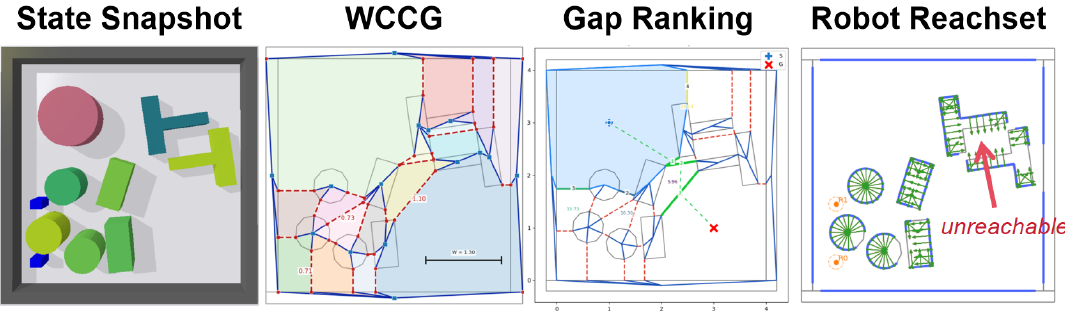}
  \vspace{-2mm}
  \caption{Overview of the proposed physics-informed hybrid search (PIHS):
  WCCG construction, gap ranking, mode generation and physics validation.}
  \label{fig:pihs_overview}
  \vspace{-4mm}
\end{figure}

The feasibility of routing the vehicle reduces to whether a~$W$-width corridor
connects~$\mathbf{x}_\texttt{V}^{\texttt{S}}$ and~$\mathbf{x}_\texttt{V}^{\texttt{G}}$.
A W--clearance connectivity graph (WCCG) encodes obstacle adjacencies under
clearance~$W$ for fast queries, as shown in Fig.~\ref{fig:pihs_overview}.
Grid- and sampling-based clearance checks typically depend on a chosen
grid or sampling resolution and must update their queries when obstacles
move~\cite{karaman2011samplingbasedalgorithmsoptimalmotion}.
Built from obstacle geometry, the WCCG needs no grid-resolution parameter in these
repeated queries.

\subsubsection{Graph Construction}
The WCCG is built by decomposing all clearance-relevant obstacle and workspace
boundaries into convex components or segments.
From each component~$C$,
a centroid node~$v_c$ is created. When two components~$C_u$ and~$C_v$
have closest points~$p_u\in C_u$ and~$p_v\in C_v$ with distance less than~$W$,
bridge nodes are added at~$p_u$ and~$p_v$. These bridge nodes are
connected by a bridge--bridge edge, annotated with the gap width
$w_{uv}\triangleq\|p_u-p_v\|$, and further linked back to their centroids with
centroid--bridge edges. Narrow passages with~$w_{uv}<W$ are marked
as potential bottlenecks. The resulting graph is given by:
\begin{equation}\label{eq:wccg}
\mathcal{G}_W\triangleq(\mathcal{V},\; \mathcal{E}_c\cup\mathcal{E}_b),
\end{equation}
where~$\mathcal{V}$ contains centroid and bridge nodes;
$\mathcal{E}_c$ contains centroid--bridge edges; and~$\mathcal{E}_b$ contains
bridge--bridge edges annotated by gap width.

\subsubsection{Connectivity Criterion}\label{subsubsec:wccg-criterion}
Once~$\mathcal{G}_W$ has been constructed, connectivity queries can be
performed without explicitly computing a geometric path. Let
$\textsf{Face}(\mathcal{G}_W)$ denote the set of connected planar faces induced
by the embedded WCCG, i.e., the maximal regions separated by obstacle
components and WCCG bottleneck edges. The criterion for~\eqref{eq:wclear}
is given by
\begin{equation}\label{eq:wccg_criterion}
\big(\mathbf{x}_\texttt{V}^{\texttt{S}},\mathbf{x}_\texttt{V}^{\texttt{G}}\in
\mathcal{F}_W(t)\big)
\wedge
\big( \exists F\in\textsf{Face}(\mathcal{G}_W):
{\mathbf{x}_\texttt{V}^{\texttt{S}},\mathbf{x}_\texttt{V}^{\texttt{G}}}\in F \big),
\end{equation}
where the faces follow the standard configuration-space inflation
interpretation: expanding obstacles by a disk of radius $W/2$ turns gaps
narrower than $W$ into overlapping barriers, and non-convex obstacles are decomposed into clearance-relevant components, after which WCCG records the separating bottlenecks. The consistency of WCCG connectivity with both continuous and rasterized configuration-space references is further evaluated numerically
in Sec.~\ref{subsec:wccg-validation}.
A frontier-tracing procedure, similar to the
BugPlanner~\cite{McGuireCroonTuyls2019}, is then used to extract a skeleton
$\Sigma$ as an ordered sequence of centroid and bridge nodes. Starting from
$\mathbf{x}_\texttt{V}^{\texttt{S}}$, it casts a ray toward
$\mathbf{x}_\texttt{V}^{\texttt{G}}$ and follows the encountered frontier loop
until either a valid exit is found or a blocking cycle is returned.

\subsection{Ranking of Frontier Gaps}\label{subsec:gap}

When \eqref{eq:wccg_criterion} fails, the planner ranks blocking gaps on
the reachable frontier of~$\mathcal{G}_W$ by predicted downstream cost and
passes them to the hybrid search as local clearance targets, linking the gap-ranking
and pushing modules in Fig.~\ref{fig:overall}.

\subsubsection{Frontier Extraction and Candidate Gaps}
A BugPlanner query on~$\mathcal{G}_W$ above
either confirms connectivity or returns a counter--clockwise frontier loop
$\mathcal{L}$ that separates the two endpoints. The bridge--bridge edges on
$\mathcal{L}$ define the first--hop candidate set
$\Gamma_\mathcal{L}\triangleq\{g_1,\cdots,g_K\}$,
where each gap~$g_k$ is represented by its two closest boundary points, its
width~$w_{g_k}<W$, and the adjacent obstacles that can potentially be moved.

\subsubsection{Evaluation for Immediate Cost}
Each candidate~$g\in\Gamma_\mathcal{L}$ is evaluated by combining the
cost for the robots to reach the gap and the effort required to widen it.
Let~$\mathbf{s}_{\mathcal{R}}$ be the current robot positions,
and~$\mathbf{o}_g$ the outside insertion point of gap~$g$. The
resulting one--hop cost is given by:
\begin{equation}\label{eq:step_cost}
\mathsf{C}\big(g \mid \mathcal{L}, \mathbf{s}_{\mathcal{R}}\big)
=\lambda_\texttt{t}\,\mathsf{C}_\texttt{t}\!\left(\mathbf{s}_{\mathcal{R}},\mathbf{o}_g\right)
+\lambda_\texttt{p}\,\big( \mathsf{C}_\texttt{dp}(g) + \mathsf{C}_\texttt{rp}(g)\big),
\end{equation}
where~$\lambda_\texttt{t},\lambda_\texttt{p}>0$ weight the costs and $\mathsf{C}_\texttt{t}$, $\mathsf{C}_\texttt{dp}$, $\mathsf{C}_\texttt{rp}$ are the transition, direct and recursive terms.

\begin{figure}[t!]
  \centering
  \includegraphics[width=0.90\linewidth]{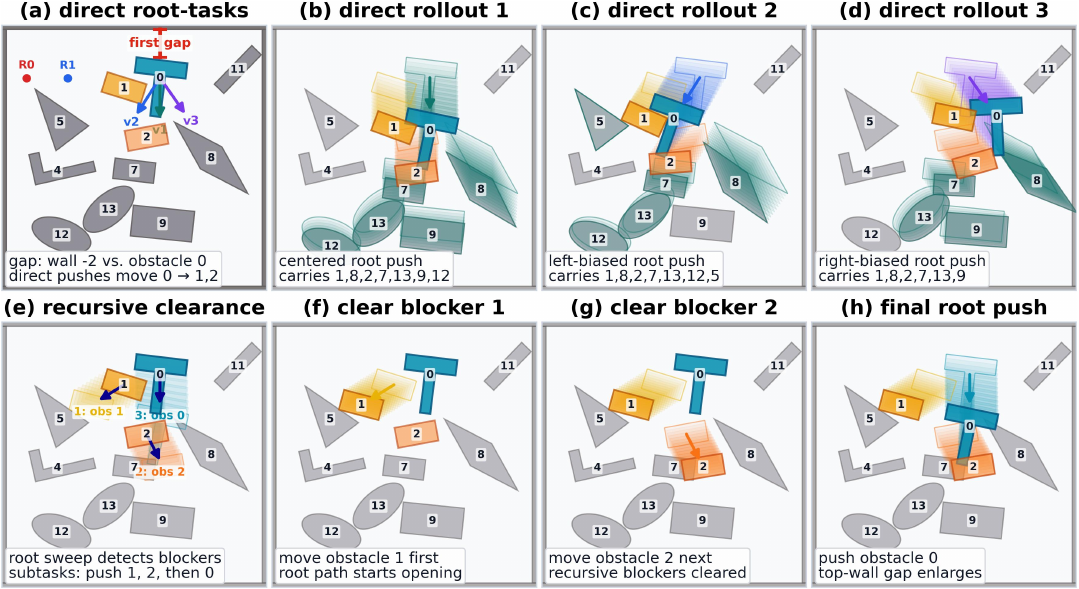}
  \vspace{-2mm}
  \caption{\textbf{Direct pushing (Top)} moves a gap-side blocker along a gap-opening
  direction, whereas \textbf{recursive pushing (Bottom)} first clears movable
  blockers in the swept region and then returns to the root gap-opening task.}
  \label{fig:direct_recursive_push}
  \vspace{-4mm}
\end{figure}

\subsubsection{Long-term Cost w.r.t. Goal}
Furthermore, to prioritize gaps closer to the goal, a heuristic is added to the evaluation,
i.e.,
$h(g)=\eta\,\big\|\mathbf{o}_g-\mathbf{x}_\texttt{V}^{\texttt{G}}\big\|$,
with the scaling constant~$\eta>0$.
The \emph{gap-lookahead} procedure performs an A$^\star$-style presearch with
bounded width and depth, virtually crossing gaps and recomputing frontiers.
The \emph{cost-to-connect} is predicted by:
\begin{equation}\label{eq:pred_cost}
\widehat{\mathsf{Cost}}(g)\triangleq
\mathsf{C}\!\big(g \mid \mathcal{L},\mathbf{s}_{\mathcal{R}}\big)
+\!\!\!\sum_{g'\in\Pi^\star(g)}\!\!\!\big{\{}\mathsf{C}(g' \mid \cdot)+h(g')\big{\}},
\end{equation}
where~$\Pi^\star(g)$ is the sequence of subsequent gaps discovered after virtually crossing~$g$. The module returns the ranked list
$\textsf{Rank}\big(\Gamma_\mathcal{L}\big)
\triangleq \textbf{argsort}_{g\in\Gamma_\mathcal{L}}\widehat{\mathsf{Cost}}(g)$,
in ascending predicted cost, so promising gaps are expanded first.
This geometric presearch estimates a promising bottleneck sequence without
moving obstacles; the associated pushing actions are generated and evaluated
in Sec.~\ref{subsec:push}.

\subsection{Pushing Mode Generation and Evaluation}\label{subsec:push}
For a selected frontier gap~$g$, this subsection constructs candidate
pushing strategies
$\Xi_g\triangleq\{(\Omega_m,\mathbf{v},\boldsymbol{\xi})\}$,
where $\mathbf{v}=(v_x,v_y,\omega)\in\mathbb{R}^3$ is a short-horizon pushing
direction for~$\Omega_m$, and $\boldsymbol{\xi}=(\mathbf{c}_m,\mathbf{u})$
is its mode in~\eqref{eq:transition}.
Wrench feasibility screens contact configurations that can realize
the required motion within actuation limits. The forces certify feasibility
rather than serving as hardware setpoints; candidate motion is then checked
by rollout or quick-pass approximation.

\subsubsection{Direct and Recursive Pushing Tasks}
For \emph{direct push}, let~$\Omega_m$ denote one of the two movable obstacles adjacent to~$g$. For
each such obstacle, a small set of candidate body velocities
$\{\mathbf{v}_m^k\}_{k=1}^{K_m}$ is sampled with a bias toward enlarging the
gap~$g$, while a few angular components are included to account for local
geometry. If the swept region of a direct push intersects movable blockers,
\emph{recursive push} tasks are required instead: for each out-layer
blocker~$\Omega_m$, the nominal direction moves it away from both the swept
region and the reference gap sequence~$\Pi^\star(g)$. The bounded dependency chain in Fig.~\ref{fig:direct_recursive_push}
prioritizes the outermost blocker~$\Omega_m$.
Clearing this blocker can free space for its parent push, allowing the search
to return toward the original gap-opening task.

\subsubsection{Pushing Mode Generation}
\begin{figure}[t!]
  \centering
  \includegraphics[width=0.90\linewidth]{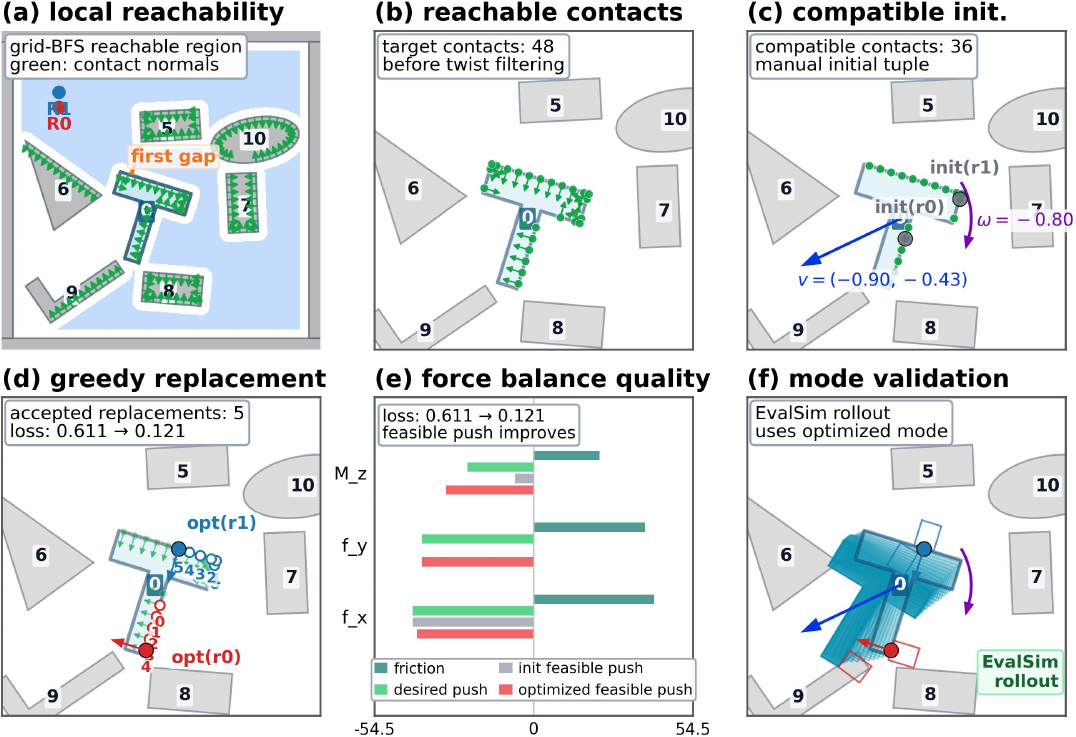}
  \vspace{-2mm}
  \caption{Mode generation for a gap-opening task: (a) local reachability,
  (b) the reachable contact candidates after twist filtering, (c) a compatible
  initial contact tuple, (d) greedy replacement of individual contacts, (e) the
  force balance of the optimized push, and (f) rollout validation of the mode.}
    \label{fig:mode_generation}
   \vspace{-4mm}
\end{figure}

At a search state~$\mathbf{s}$, a pushing mode is generated for each candidate
direction~$\mathbf{v}_m^k$ and obstacle~$\Omega_m$. In this subsubsection,
contact tuples, object velocities, forces, and wrenches are expressed in the
body frame of~$\Omega_m$. For a contact tuple~$\mathbf{c}$, its
multi-directional force feasibility is:
\begin{equation}\label{eq:mode-feasibility}
\mathrm{J}_{\texttt{MF}}^m(\mathbf{c},\mathbf{v}_m^k)
\triangleq
\sum_{\mathbf{v}_d\in\mathcal{D}_m^k} w_d
\underset{\mathbf{u}\in{U}_m(\mathbf{c})}{\textbf{min}}
\|
\mathbf{q}_m(\mathbf{c},\mathbf{u})
+\mathbf{q}_m^\mu(\mathbf{v}_d)
\| ,
\end{equation}
where~$\|\cdot\|$ is a weighted $\ell_1$ wrench norm, the finite set
$\mathcal{D}_m^k$ contains perturbations of~$\mathbf{v}_m^k$ with weights $w_d>0$, and
$\mathbf{q}_m^\mu(\mathbf{v}_d)$ is the quasi-static friction wrench; the inner
minimization is a small LP over the admissible forces ${U}_m(\mathbf{c})$.
A state- and direction-dependent contact set
$\mathcal{A}_m^k(\mathbf{s})\subset\partial\Omega_m$ is drawn from the
robot-reachable boundary of~$\Omega_m$ by velocity--force compatibility
filtering (Fig.~\ref{fig:mode_generation}), retaining only contacts with collision-free pre-contact poses. Starting from cached layouts or
random initializations, sequential greedy replacement refines the tuple.
At iteration~$\ell$, $\mathbf{c}_{r,c}^{\ell}$ denotes the tuple with contact~$r$
replaced by candidate~$c$. The replacement is:
\begin{equation}
c^\star
=
\underset{c\in\mathcal{A}_m^k(\mathbf{s})}{\textbf{argmin}}\;
\mathrm{J}_{\texttt{MF}}^m
\bigl(\mathbf{c}_{r,c}^{\ell},\mathbf{v}_m^k\bigr).
\label{eq:greedy_mode_update}
\end{equation}
The tuple is updated with $c^\star$. Contact slots are swept until
$\mathrm{J}_{\texttt{MF}}^m$ stops decreasing or the evaluation budget is exhausted.
The best tuple over initializations gives
$\mathbf{c}_m^{k,\star}$, from which $\mathbf{u}_m^k$ follows from the inner
minimization in~\eqref{eq:mode-feasibility}, yielding
$\boldsymbol{\xi}_m^k=(\mathbf{c}_m^{k,\star},\mathbf{u}_m^k)$; each mode is
stored in the \emph{mode-cache} for subsequent warm starts. If the sequential replacement cannot assign a
contact to every robot, a bounded backtracking search over robot--contact
combinations attempts the full-team assignment. If no feasible full-team mode is found
within this budget, for instance on small faces or under the pairwise
pusher-clearance constraint, one robot is temporarily left inactive and the
assignment is repeated with the remaining team. Reduced-team modes
carry a ranking penalty, so a feasible full-team mode is preferred, and the
inactive robot returns for the next task. The best few admissible pairs are retained for gap~$g$.

\subsubsection{Pushing Strategy Evaluation}

For a current state~$\mathbf{s}$ and candidate strategy
$(\Omega_m,\mathbf{v},\boldsymbol{\xi})\in\Xi_g$, a short-horizon rollout
evaluates the transition in~\eqref{eq:transition} under the selected mode.
As shown in Fig.~\ref{fig:mode_generation}, a strategy is accepted if the
robots reach the pre-contact poses, maintain stable contact and induce
obstacle motion above threshold. The simulation returns:
\begin{equation}
    (\mathbf{s}',\,\delta_{\texttt{T}},\,\delta_{\texttt{J}})
    \triangleq
    \mathsf{EvalSim}
    \big(
    \mathbf{s},(\Omega_m,\mathbf{v},\boldsymbol{\xi})
    \big),
    \label{eq:sim}
\end{equation}
where $\mathbf{s}'$ contains the resulting robot and object configurations,
$\delta_{\texttt{T}}>0$ is duration and $\delta_{\texttt{J}}$ control effort.
Joint rollout captures contact-induced interactions among all bodies.
Optionally, quick-pass approximates a transition using a cached rollout
or a mechanically feasible reference trajectory, subject to compatible relative
configurations and physical parameters, reachability, and swept-clearance checks.
The trajectory endpoint and associated duration and effort estimates supply
$\mathbf{s}'$, $\delta_{\texttt{T}}$, and $\delta_{\texttt{J}}$.
This shortcut trades prediction accuracy for reduced online simulation effort.
\begin{figure}[t!]
  \centering
  \includegraphics[width=1.0\linewidth]{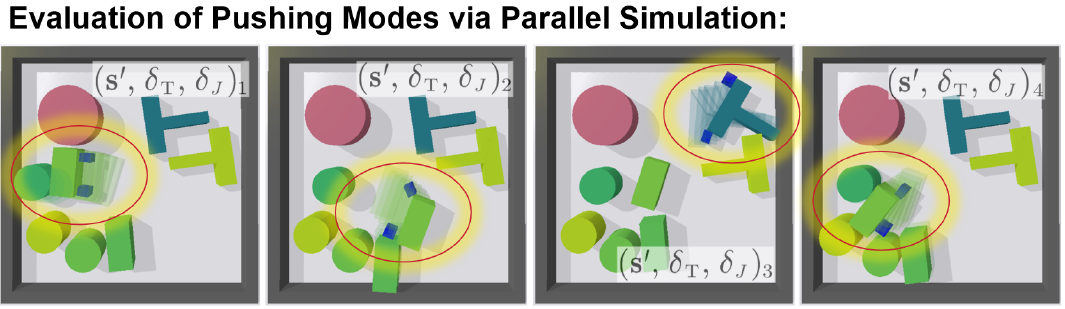}
  \vspace{-6mm}
  \caption{Sampled push tasks are evaluated via parallel simulation,
  producing updated states for search expansion.}
    \label{fig:PIHS}
   \vspace{-5mm}
\end{figure}

\subsection{Physics-Informed Hybrid Search}\label{subsec:simloop}

The physics-informed hybrid search (PIHS) selects blocking gaps and
evaluates pushing strategies in parallel
(Fig.~\ref{fig:PIHS}, Alg.~\ref{alg:pihs}), using short-horizon simulation or
the optional quick-pass approximation of Sec.~\ref{subsec:push}.
The evaluated clearance cost of a partial schedule in~\eqref{eq:problem}
supplies the accumulated-cost term in the node priority
of~\eqref{eq:priority}.
Finite sampling gives no global optimality or completeness guarantee.
\subsubsection{Tree and Initialization}
Tree nodes $\nu=(\mathbf{s},\pi)$ contain the states in~\eqref{eq:transition}
and partial schedule in~\eqref{eq:schedule}.
$\mathsf{Rank}(\nu)$ orders frontier gaps using \eqref{eq:pred_cost}, and
$\chi(\nu)$ accumulates execution cost from the root.
Alg.~\ref{alg:pihs} initializes $\nu_0=(\mathbf{s}_0,\emptyset)$ and its queue.

\subsubsection{Node Selection}
At each iteration, line~\ref{algline:select} of Alg.~\ref{alg:pihs} selects
the node with minimum priority:
\begin{equation}\label{eq:priority}
  f(\nu)\triangleq \chi(\nu)+\underset{g\in \textsf{Rank}(\nu)}{\textbf{min}}
  \widehat{\mathsf{Cost}}(g).
\end{equation}
The second term estimates the minimum remaining cost over unexplored gaps.
Gap ranking orders local clearance targets within a node, whereas this
priority compares partial schedules with different clearing histories.
This heuristic need not be admissible, so the priority gives no A* optimality
guarantee.

\begin{figure*}[t!]
  \centering
  \includegraphics[width=0.95\linewidth]{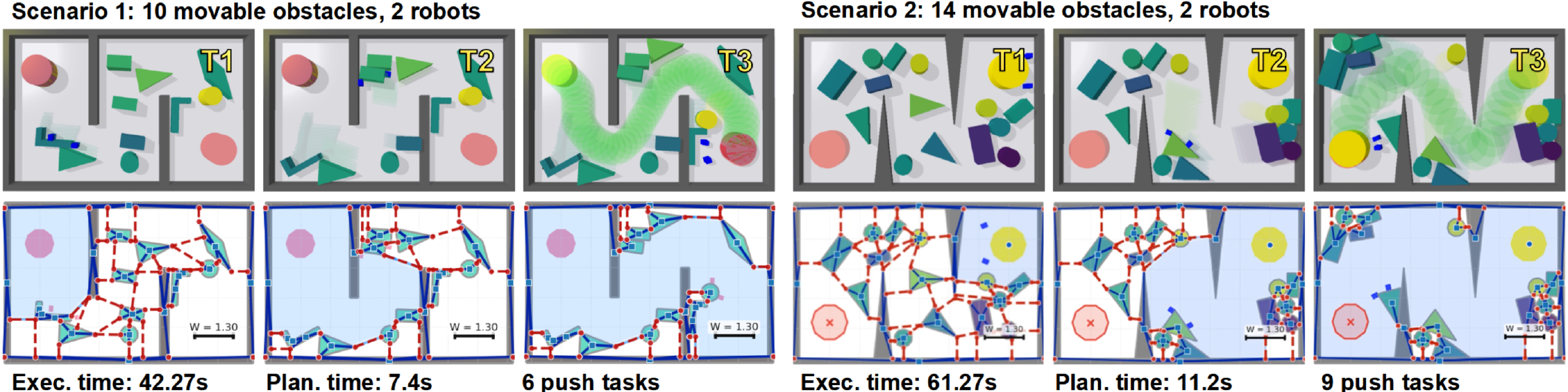}
  \vspace{-2mm}
  \caption{Execution and planning alignment in Scenario 1 (\textbf{Left})
    and  Scenario 2 (\textbf{Right}).
\textbf{Top:} simulated snapshots during path clearing.
\textbf{Bottom:} corresponding WCCG overlays with the start face in blue.
The executed pushes progressively expand the reachable $W$--clearance region until the goal is connected.}
\vspace{-5mm}
\label{fig:main_result}
\end{figure*}

\subsubsection{Node Expansion with Parallel Simulation}
Alg.~\ref{alg:pihs} expands the selected node~$\nu=(\mathbf{s},\pi)$ in
lines~\ref{algline:expand}--\ref{algline:reinsert}. For each sampled gap
$g \in \mathsf{Rank}(\nu)$,
its predicted gap sequence~$\Pi^\star(g)$ in~\eqref{eq:pred_cost}
is first derived. Then, fresh direction samples and mode optimization by~\eqref{eq:mode-feasibility}
provide a candidate strategy set
$\Xi_g \triangleq \{(\Omega_m,\mathbf{v},\boldsymbol{\xi})\}$,
including direct pushes of gap-side obstacles and recursive pushes of outer
blockers.
Thus, expansion is prioritized first by the gap ordering in
$\mathsf{Rank}(\nu)$, and then by the candidate ordering within~$\Xi_g$.
Each candidate is evaluated by~\eqref{eq:sim}, using quick-pass approximation
or a rollout as described in Sec.~\ref{subsec:push}.
A successful evaluation yields
$(\mathbf{s}',\delta_{\texttt{T}},\delta_{\texttt{J}})$, from which a child
node~$\nu' \triangleq (\mathbf{s}',\pi')$ is created with
$\pi' \triangleq \pi \cup \{\tau\}$, appending
$\tau \triangleq (\Omega_m,\mathbf{v},\boldsymbol{\xi})$ to~$\pi$.
The cumulative cost is updated by:
$\chi(\nu') \triangleq \chi(\nu) + \widehat{\mathsf{C}}(\nu,\tau)$,
where~$\widehat{\mathsf{C}}(\nu,\tau)\triangleq
\delta_{\texttt{T}} + \alpha\,\delta_{\texttt{J}}$
is the incremental cost.
With \emph{reinsertion} in line~\ref{algline:reinsert}, the parent node is requeued
for later expansion with new target and direction samples.

{
\SetAlFnt{\small}
\SetAlgoSkip{0pt}
\begin{algorithm}[t!]
\caption{Physics-Informed Hybrid Search}\label{alg:pihs}
\SetAlgoLined
\KwIn{$\mathbf{s}_0$,~$\alpha$,~$\mathsf{EvalSim}(\cdot)$,~$W$--criterion by~\eqref{eq:wccg_criterion}}
\KwOut{Clearing schedule $\pi^\star$, or failure}
$\nu_0\gets(\mathbf{s}_0,\emptyset),\;\chi(\nu_0)=0$\;
Init~$\mathcal{Q}$ by~\eqref{eq:priority}\;
\While{$\mathcal{Q}\ne\emptyset$ and budget remains}{
 $\nu=(\mathbf{s},\pi)\gets\mathcal{Q}.\texttt{pop\_min}()$ by~\eqref{eq:priority}\nllabel{algline:select}\;
 \If{$\mathbf{s}$ admits~$\mathcal{P}^W_{\texttt{V}}$ by~\eqref{eq:wccg_criterion}}{
  \Return $\pi$\;
 }
  \ForEach{sampled $g\in\mathsf{Rank}(\nu)$\nllabel{algline:expand}}{
    Compute~$\Pi^\star(g)$, generate~$\Xi_g$\;
    \ForEach{$(\Omega_m,\mathbf{v},\boldsymbol{\xi})\in\Xi_g$}{
      $\tau=(\Omega_m,\mathbf{v},\boldsymbol{\xi})$ \tcp{Direct or recur.}
      $(\mathbf{s}',\delta_\texttt{T},\delta_\texttt{J})\gets\mathsf{EvalSim}(\mathbf{s},\tau)$
      by~\eqref{eq:sim}\;
    \If{success}{
     $\pi'\gets\pi\cup\{\tau\}$\;
     $\nu'\gets(\mathbf{s}',\pi')$\;
     $\chi(\nu')\gets\chi(\nu)+\delta_\texttt{T}+\alpha\,\delta_\texttt{J}$\;
     Add $\nu'$ to $\mathcal{Q}$\nllabel{algline:add}\;
      }
    }
  }
 Reinsert $\nu$ into $\mathcal{Q}$ for further sampling\nllabel{algline:reinsert}\;
}
\Return failure;
\end{algorithm}
}

\subsubsection{Termination}
Alg.~\ref{alg:pihs} terminates when a node admits a $W$--clear path by
\eqref{eq:wccg_criterion}. The returned schedule satisfies nominal clearance
and evaluated pushing transitions but need not minimize
$\mathcal{J}_{\mathrm{clr}}$ in~\eqref{eq:problem}.
Vehicle traversal follows clearing.

\begin{remark}\label{remark:simloop}
Gap-lookahead reduces unproductive branches, \textsc{mode-cache}
warm-starts optimization, and optional quick-pass approximates transitions
to reduce rollout effort.
\hfill$\blacksquare$
\end{remark}

\subsection{Overall Analyses}\label{subsec:overall}

\subsubsection{Execution and Online Adaptation}\label{subsec:execute}
The schedule $\pi=\{\tau_1,\cdots,\tau_K\}$ is executed in a receding-horizon
manner, with $\tau_k=(\Omega_{m_k},\mathbf{v}_k,\boldsymbol{\xi}_k)$.
Robots reach the assigned contacts and follow $\mathbf{v}_k$ from the current
state in~\eqref{eq:sim}. On hardware, the optimized forces certify wrench
feasibility within actuation limits; they are not control setpoints.
At each control interval, $\mathsf{EvalSim}$ predicts object motion tracked with
motion-capture feedback. Between pushes, robot reassignment preserves contact
order along the obstacle boundary~\cite{tang2026pushingbots} to reduce path crossings.
After each push, the state and WCCG are updated until a $W$-clear path connects
$\mathbf{x}_\texttt{V}^{\texttt{S}}$ and $\mathbf{x}_\texttt{V}^{\texttt{G}}$.
Failed contacts, stalled pushing, or invalidated connectivity trigger replanning
by Alg.~\ref{alg:pihs} from the latest observed state.

\subsubsection{Computational Complexity}\label{subsubsec:complexity}

Given~$M$ movable obstacles, the computational cost of one node
expansion in Alg.~\ref{alg:pihs} is dominated by candidate generation and
physics validation. WCCG updating and gap ranking on the node's frontier loop~$\mathcal{L}_\nu$ incur
$\mathcal{O}(M\log M+|\mathcal{V}|+|\mathcal{E}_c|+|\mathcal{E}_b|)$ and
$\mathcal{O}(|\Gamma_{\mathcal{L}_\nu}|
\log|\Gamma_{\mathcal{L}_\nu}|)$ costs.
The dominant cost is physics validation, which grows with the number of
evaluated candidates; parallel workers reduce the rollout wall time.
Robot-team size affects mode generation: each greedy sweep evaluates at most $N|\mathcal{A}|$ single-contact
substitutions, where $|\mathcal{A}|$ bounds the reachable contact candidates considered per
robot, rather than $|\mathcal{A}|^N$ joint tuples, and refinement is capped by a bounded evaluation budget; failures are handled by bounded backtracking with budget $B$ and a reduced-team fallback over
at most $N$ one-robot-excluded subsets. Each candidate is checked by an LP over at most $2N$ force variables.

\section{Experiments}
\label{sec:experiments}
PIHS is evaluated in simulation and hardware for connectivity, clearing
performance, scalability, and robustness to execution-side physical mismatch.
The algorithms use \texttt{Python3} and \texttt{PyBullet}~\cite{coumans2019}
on an Intel i9-14900KF with an RTX 4080.
The accompanying \href{https://youtu.be/HkQIcVWiIuQ}{video} illustrates the simulation and hardware experiments.
\subsection{Numerical Simulations}
\label{subsec:sim}
\begin{table}[t]

\centering

\caption{WCCG connectivity validation and mean runtime.}

\label{tab:wccg-validation}

\vspace{-2.0mm}

\footnotesize

\setlength{\tabcolsep}{3.8pt}
\setlength{\belowrulesep}{0.30ex}
\renewcommand{\arraystretch}{0.86}

\begin{tabular}{
 @{}
 l
 cc
 @{\hspace{8pt}}
 ccc
 @{}
}

\toprule

&
\multicolumn{2}{c}{Validation}
&
\multicolumn{3}{c}{Runtime (ms)}
\\[-0.6ex]

\cmidrule(lr){2-3}
\cmidrule(lr){4-6}

Type
& CC $=$ FR
& FP/FN$^\dagger$
& CC
& FR
& WCCG
\\

\midrule

Convex
& 100/100
& 0/0
& 1.23
& 282.9
& \textbf{0.86}
\\

Non-convex
& 98/100
& 0/0
& 1.51
& 204.1
& \textbf{1.30}
\\

Mixed
& 98/100
& 0/0
& 1.33
& 242.4
& \textbf{1.05}
\\

\midrule

Total
& 296/300
& \textbf{0/0}
& 1.36
& 243.1
& \textbf{1.07}
\\

\bottomrule

\end{tabular}

\vspace{0.5mm}

{\scriptsize
$^\dagger$WCCG errors relative to the continuous configuration-space reference (CC).
All runtimes use the same 300-scene harness with the compiled WCCG backend.
}

\vspace{-5.5mm}

\end{table}

\subsubsection{Setup and Configuration}\label{subsec:sim-setup}

The simulation step is $\Delta t=1/240$\,s and the control period is $1/40$\,s.
Robots are disk or box pushers, and obstacle masses range from $5$ to $15$\,kg
in Scenarios~1--2 and from $20$ to $30$\,kg in the generalization studies.
The nominal $8\times5$\,m workspace contains bottleneck-forming bar obstacles,
with two robots and $10$--$15$ movables under minimum-separation constraints.
Unless stated otherwise, comparisons and ablations use Scenario~1 with ten
movables and $40$ trials; Scenario~2 has $14$ movables and provides an additional
cluttered case in Fig.~\ref{fig:main_result}. All methods share robot footprints,
obstacle configurations, collision and reachability checks, and the execution
simulator. They use eight workers, an $800$\,s planning cap, and an execution cap
of $20{,}000$ control steps, equivalent to $500$\,s. A trial succeeds when the
vehicle reaches its goal through a $W$-clear path. Planning time $PT$ and execution time $ET$ average over successful
trials unless stated otherwise; failures count only in success rates, while
scalability timing curves cap failures at the budgets.

\begin{table}[t]
 \centering
 \begin{threeparttable}
 \caption{Performance comparison and ablations on Scenario~1 over $40$ randomized trials.\tnote{a}}
 \label{tab:main_ablation}
 \setlength{\tabcolsep}{2.7pt}
 \renewcommand{\arraystretch}{0.92}
 \begin{tabular}{lccccc}
 \toprule
 Method / Variant & Succ.(\%) & PT (s) & ET (s) & \#Sims & \#Pushes \\
 \midrule
 DFS-WCCG & 25.0 & $>$100.0 & 51.2 & $>$500 & 8.0 \\
 SL-Push (offline) & 62.5 & 0.06 & 44.1 & 0.0 & 7.3 \\
 SL-Push (sim) & 75.0 & 18.2 & 64.6 & 20.0 & 10.2 \\
 Rec-NAMO & 37.5 & 13.3 & 42.4 & 0.0 & 7.8 \\
 \textbf{PushAround (ours)}
 & \textbf{97.5} & \textbf{10.3} & \textbf{28.6}
 & \textbf{121.5} & \textbf{6.0} \\
 \midrule
 w/o gap-lookahead & 82.5 & 16.9 & 35.2 & 178.3 & 7.2 \\
 w/o mode-cache & 92.5 & 17.1 & 33.4 & 164.6 & 6.5 \\
 w/o quick-pass & 90.0 & 23.6 & 41.9 & 208.2 & 6.1 \\
 w/o reinsertion & 85.0 & 12.8 & 31.1 & 151.7 & 7.8 \\
 \bottomrule
 \end{tabular}
 \begin{tablenotes}\scriptsize
 \item[a] Times are averaged over successful trials; failures count in the
 success rate only. For DFS-WCCG, $PT$ and
 \#Sims are not comparable and are given as lower bounds ($>$).
 \end{tablenotes}
 \end{threeparttable}
 \vspace{-4.0mm}
\end{table}

\begin{figure}[t!]
 \centering
 \includegraphics[width=0.8\columnwidth]{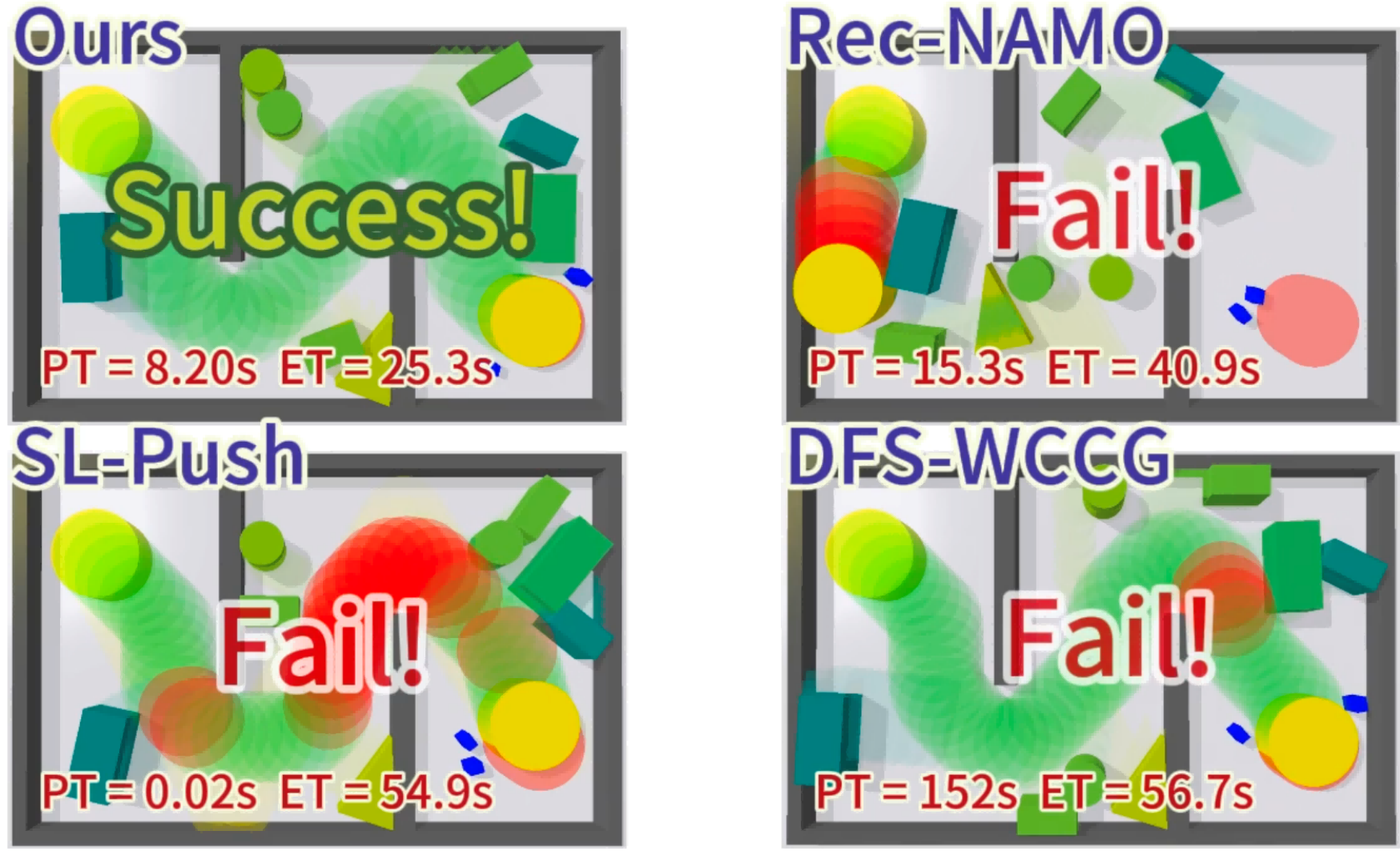}
 \vspace{-2.0mm}
 \caption{Failure cases by baselines Rec-NAMO, SL-Push and DFS-WCCG.}
 \label{fig:baseline}
 \vspace{-6.0mm}
\end{figure}

\subsubsection{Results}
Across the $40$ Scenario~1 trials, PushAround achieves a $97.5\%$ success rate,
with mean planning and execution times of $10.3$\,s and $28.6$\,s, respectively,
and $6.0$ pushes per successful trial, as shown in Table~\ref{tab:main_ablation}.
Fig.~\ref{fig:main_result} shows representative runs in both scenarios.
In Scenario~1 with $10$ movable obstacles and two robots, PushAround generates
a $6$-push plan in $7.4$\,s and completes execution in $42.3$\,s. In
Scenario~2 with $14$ obstacles, it generates a $9$-push plan in $11.2$\,s,
with $61.3$\,s of execution.
Planning effort is dominated by simulation-based validation: a successful
trial uses $121.5$ simulations on average for $6.0$ executed pushes. The
ablations indicate that mode-cache, quick-pass, gap-lookahead, and reinsertion
mainly affect this validation load rather than plan length. Execution takes
longer because each selected push is physically carried out and observed,
whereas inter-push repositioning is handled by geometric reachability checks.

\subsubsection{WCCG Connectivity Validation}
\label{subsec:wccg-validation}

Table~\ref{tab:wccg-validation} compares WCCG against two references on $300$
random convex, non-convex, and mixed scenes. The continuous reference (CC) resolves collisions geometrically after inflating boundaries by~$W/2$, with no grid discretization~\cite{lozano1983spatial}; the fine-raster reference (FR) discretizes the workspace on a $0.01$\,m occupancy grid. CC and FR disagree in $4$ of $300$ cases, all near narrow contacts and rasterization-related, whereas WCCG matches CC on all $300$ scenes without false positives or negatives. A query takes $1.07$\,ms, comprising $0.67$\,ms for querying and $0.40$\,ms for
construction, versus $1.36$\,ms for CC and $243$\,ms for FR. It also supplies the gap graph for PIHS. Per-push WCCG rebuilding is negligible next to physics rollouts.

\begin{table}[t]

\caption{Scalability and robustness.}
\label{tab:generalization}
\vspace{-2.0mm}
\centering
\footnotesize

\renewcommand{\arraystretch}{0.95}
\setlength{\tabcolsep}{3.5pt}

\begin{tabular}{
 @{}
 l
 c
 c
 r@{\;}c@{\;}l
 @{\hspace{10pt}}
 r@{\;}c@{\;}l
 @{}
}
\hline

Case
& Trials
& Succ.(\%)
& \multicolumn{3}{c@{\hspace{10pt}}}{$PT$ (s)}
& \multicolumn{3}{c}{$ET$ (s)}
\\

\hline

\multicolumn{9}{l}{
 \emph{Obstacle density} (PushAround, $N=2$, nominal)
}
\\

$M=15$
& 10 & 100
& 2.5 & $\pm$ & 0.9
& 37.1 & $\pm$ & 21.0
\\

$M=30$
& 10 & 100
& 11.7 & $\pm$ & 3.7
& 100.9 & $\pm$ & 34.7
\\

$M=45$
& 10 & 100
& 53.4 & $\pm$ & 24.5
& 272.7 & $\pm$ & 97.4
\\

\hline

\multicolumn{9}{l}{
 \emph{Robot-team size} ($M=30$, nominal; 10 trials per cell)
}
\\

PushAround, $N=3$
& 10 & 100
& 30.0 & $\pm$ & 12.7
& 97.3 & $\pm$ & 42.1
\\

PushAround, $N=4$
& 10 & 100
& 29.2 & $\pm$ & 8.9
& 110.1 & $\pm$ & 33.1
\\

SL-Push (sim), $N=3$
& 10 & 80
& 30.0 & $\pm$ & 46.3
& 193.3 & $\pm$ & 111.3
\\

SL-Push (sim), $N=4$
& 10 & 80
& 23.4 & $\pm$ & 18.2
& 171.7 & $\pm$ & 61.2
\\

DFS-WCCG, $N=3$
& 10 & 100
& 142.6 & $\pm$ & 183.1
& 178.1 & $\pm$ & 72.1
\\

DFS-WCCG, $N=4$
& 10 & 90
& 101.9 & $\pm$ & 103.9
& 175.9 & $\pm$ & 113.6
\\

\hline

\multicolumn{9}{l}{
 \emph{Physics mismatch} ($M=30$, $N=2$)
}
\\

Nominal
& 10 & 100
& 16.8 & $\pm$ & 3.4
& 104.7 & $\pm$ & 37.4
\\

Mass $\times 0.6$--$1.4$
& 40 & 100
& 16.8 & $\pm$ & 5.4
& 105.9 & $\pm$ & 44.6
\\

Lateral fric. $\times 0.6$--$1.4$
& 40 & 97.5
& 16.5 & $\pm$ & 5.7
& 114.0 & $\pm$ & 48.5
\\

Spin fric. $\times 0.6$--$1.4$
& 40 & 97.5
& 15.2 & $\pm$ & 3.6
& 100.1 & $\pm$ & 35.7
\\

Force limit $\times 0.6$--$1.4$
& 40 & 95.0
& 18.9 & $\pm$ & 10.9
& 115.9 & $\pm$ & 58.5
\\

All perturbed
& 160 & 97.5
& 16.9 & $\pm$ & 7.0
& 108.9 & $\pm$ & 47.4
\\

\hline
\end{tabular}
\begin{flushleft}
\footnotesize
Each perturbed row scales one parameter; the others stay nominal. Times and budget
are as in Table~\ref{tab:main_ablation}.
\end{flushleft}
\vspace{-7.0mm}
\end{table}

\subsubsection{Ablation Studies}
Four ablations remove individual mechanisms: \textbf{w/o gap-lookahead} omits
downstream costs in gap ranking; \textbf{w/o mode-cache} disables cached warm
starts; \textbf{w/o quick-pass} evaluates every candidate by rollout; and
\textbf{w/o reinsertion} expands each node only once.
Table~\ref{tab:main_ablation} shows success drops exceeding ten percentage points without
gap-lookahead or reinsertion. Removing quick-pass lowers success to $90.0\%$;
removing mode-cache increases simulations.

\subsubsection{Comparisons}\label{subsec:comparisons}
Four baseline variants run on Scenario~1 over $40$ randomized trials under
the same $W$-clearance criterion, scene distribution, and execution simulator.
\textbf{DFS-WCCG} searches fixed object--direction pushes depth first and
checks transitions by physics rollout. \textbf{SL-Push} clears blockers from a
straight corridor by pushing them sideways; its offline variant plans
geometric displacements, while its simulation-validated variant checks
candidate pushes by rollout. \textbf{Rec-NAMO} recursively clears blocking
objects using geometric planning without rollout validation during search.
These variants represent search-based, local-pushing, and recursive NAMO
strategies~\cite{stilman2007manipulation,yao2024local,saxena2023planning,ren2025search}.
Table~\ref{tab:main_ablation} shows that PushAround reaches the highest success
rate, $97.5\%$, at the lowest combined planning-and-execution time among
successful trials ($38.9$\,s).
DFS-WCCG exhibits rapid branching growth, offline SL-Push produces displacements that fail in execution, and Rec-NAMO often fails to connect the $W$-clear corridor in clutter (Fig.~\ref{fig:baseline}).

\subsubsection{Scalability and Robustness Analysis}
\label{subsec:generalization}

\begin{figure}[t!]
 \centering
 \includegraphics[width=0.9\columnwidth]{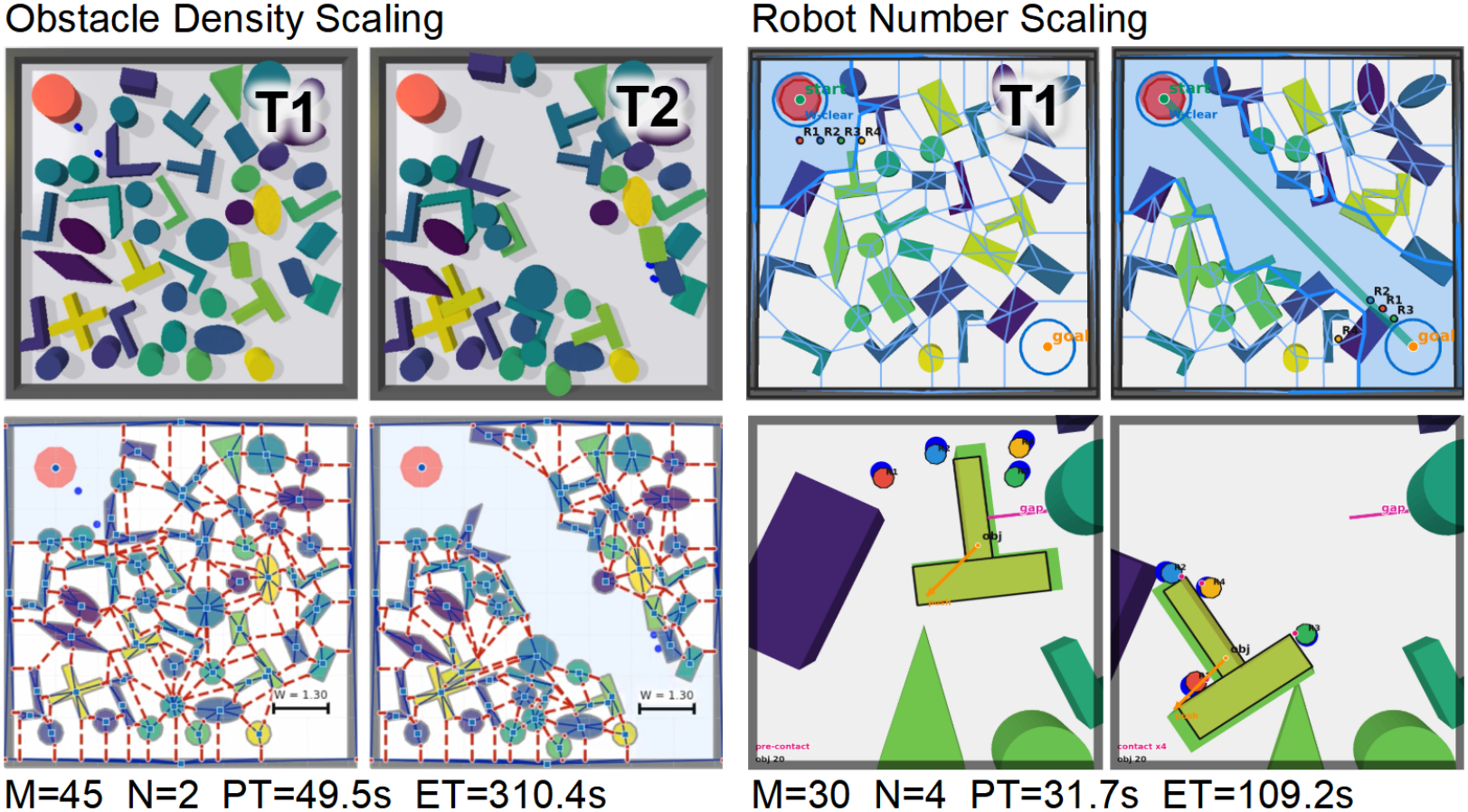}
 \vspace{-2mm}
 \caption{Scalability of the clearing process. \textbf{Left:} density scaling ($M{=}45$, two robots): snapshots during clearing (\textbf{top}) and WCCG overlays with the start face in blue (\textbf{bottom}). \textbf{Right:} team scaling ($M{=}30$, four robots): initial and final states (\textbf{top}) and one push before contact and during execution (\textbf{bottom}).}
 \label{fig:scalability_scenes}
 \vspace{-3mm}
\end{figure}

\begin{figure}[t!]
 \centering
 \includegraphics[width=0.95\columnwidth]{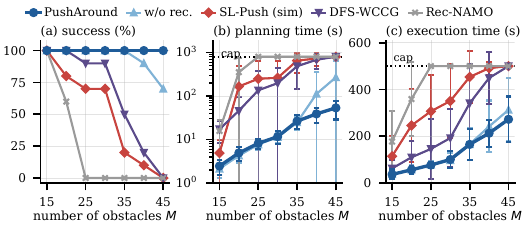}
 \vspace{-2.0mm}
 \caption{Obstacle-density scaling ($8\times 8$\,m, $10$ seeds per density). Times are
 budget-capped (planning $800$\,s, execution $20{,}000$ steps $=500$\,s; dotted).}
 \label{fig:density_scaling}
 \vspace{-5.0mm}
\end{figure}

Table~\ref{tab:generalization} and Figs.~\ref{fig:scalability_scenes}--\ref{fig:density_scaling}
report generalization in an $8\times8$\,m workspace.
Density runs query a precomputed mode table; its preparation is
excluded from $PT$. Mismatch and team-size runs optimize modes greedily.
With $M=30$ and $N=2$, nominal $PT$ is $11.7$, $16.8$, and $23.9$\,s,
respectively. The team-size study uses a different realization of the same seeds.

\textbf{(I) Obstacle density:} with two robots and ten trials per density,
PushAround maintains $100\%$ success for $M=15$--$45$, planning in
$2.5$--$53.4$\,s. At $M=45$, none of the evaluated baselines succeeds,
while the no-recursion ablation solves $7/10$ trials.
Across the clutter sweep, branching stays near six while expansions rise
from $3.5$ to $24.5$. Median search effort is ten expansions and
$48$ visited snapshots, with about $60$ simulations at $165$\,ms each versus
$22$\,ms per mode.

\textbf{(II) Robot-team size:} at $M=30$, PushAround solves all ten trials
for each of $N=2$, $3$, and $4$. The shared mode generator first tries the
full team and may temporarily leave one robot inactive.
Across $N=2$--$4$, search effort remains near ten
expansions and $60$--$63$ simulations; aggregate mode-generation time rises
from $15.8$ to $46.4$\,s.

\begin{figure*}[!t]
 \centering
 \includegraphics[width=0.95\linewidth]{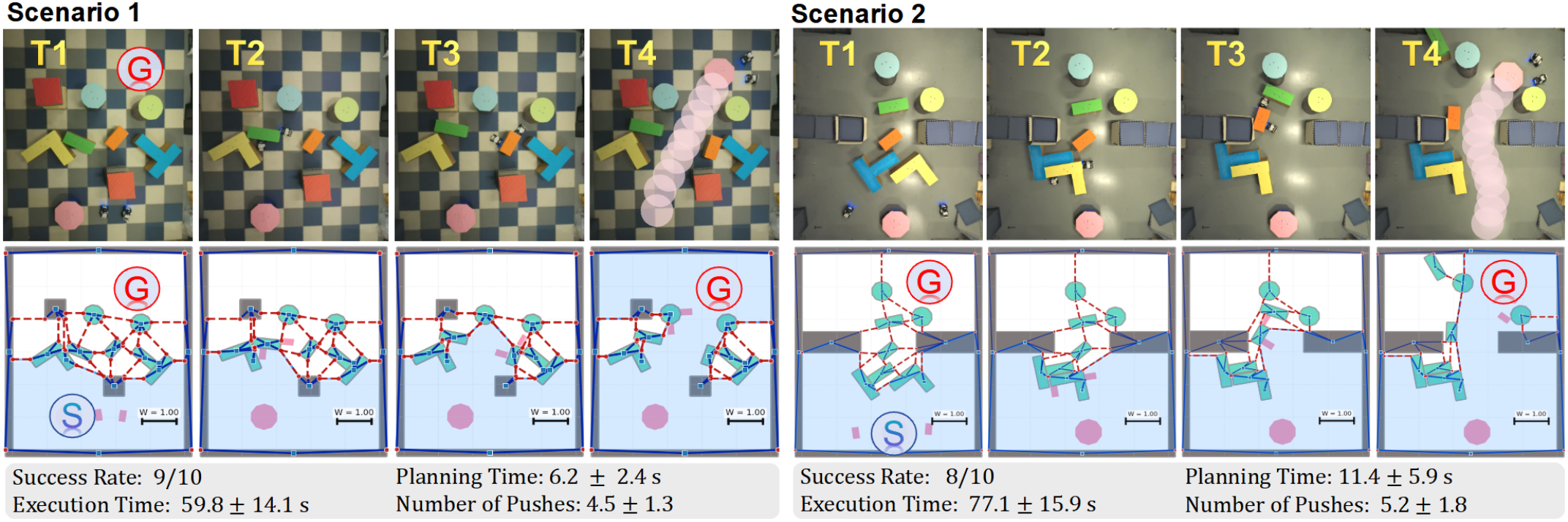}
 \vspace{-2.0mm}
 \caption{Real-world pushing experiments with execution--planning alignment (over~$20$ runs, $2$ scenarios).
 \textbf{Top:} two robots pushing obstacles in a \(5{\times}6\,\mathrm{m}\)
 workspace. \textbf{Bottom:} WCCG overlays with the start face in blue.}
 \vspace{-5.0mm}
 \label{fig:hardware}
\end{figure*}

{\clubpenalty=10000
\textbf{(III) Physics mismatch:} the planner retains nominal physics while
execution varies one parameter family at a time.
Nominal and perturbed conditions reuse ten base seeds for
PushAround and five per baseline.
Under $\pm40\%$ perturbations to four parameter families, PushAround,
SL-Push (sim), and DFS-WCCG achieve $97.5\%$, $32.5\%$, and $57.5\%$ success,
respectively. Their $PT$ averages $18.4$, $31.5$, and $229.1$\,s,
averaging successful-trial means over conditions with successes.
Nominal success is $10/10$ for PushAround and $3/5$ per baseline.
Comparisons reflect end-to-end performance, not normalized degradation;
joint errors are untested.\par}

\subsection{Hardware Experiments}\label{subsec:hardware}

\subsubsection{System Description}\label{subsec:exp-description}
Experiments use a $5\times6$\,m workspace of interlocking $0.6\times0.6$\,m foam mats (Fig.~\ref{fig:hardware}), $2$ UGVs and $6$ cardboard boxes; robots and movables carry motion-capture markers streamed to \texttt{PyBullet} in real time.
The UGVs are velocity-controlled, and the optimized contact forces are used only for feasibility evaluation, not for motor control.
Motion-capture poses are compared with $\mathsf{EvalSim}$ predictions:
position or orientation errors above $0.3$\,m or $\pi/4$ trigger replanning,
as do the failures in Sec.~\ref{subsec:execute}.

\subsubsection{Results}\label{subsec:exp-results}
Two representative real-world runs are shown in Fig.~\ref{fig:hardware}.
In hardware Scenario~1, PIHS considers five candidate gaps and plans a four-task schedule in $6.5$\,s using $12$ node expansions and $49$ visited snapshots at depth $6$; the robots then execute it, first rotating the yellow L-shaped object to open a gap near an
immovable obstacle and then separating the green and orange objects until a $W$--clear path is established at $t=52\,$s.
In hardware Scenario~2, PushAround succeeds in $8/10$ trials with an average of $5.2$
pushes and $11.4$\,s of planning. In the illustrated run, the first direct push on the blue L-shaped
object also moves the adjacent T-shaped block and opens the first gap.
Coupled rigid-body rollout in $\mathsf{EvalSim}$ predicts this neighboring-object
motion without a separately encoded action. The resulting snapshot updates
the WCCG before execution continues.
Execution takes $77.1\pm15.9$\,s (mean $\pm$ standard deviation);
online replanning handles deviations due to force limits, drift and wheel slip.

\section{Conclusion} \label{sec:conclusion}
This work presents PushAround, a multi-robot framework that jointly plans
obstacle selection, contacts and forces through a hybrid search.
Geometric gap reasoning identifies where the corridor must widen, while
pushing evaluation assesses the corresponding rearrangements under physical
constraints. Simulation studies examine obstacle density, team size, and
execution-side physics mismatch; hardware experiments demonstrate path
clearing with state-feedback replanning. Future work targets obstacle-state
estimation without external sensing or known masses.

\bibliographystyle{IEEEtran}
\bibliography{contents/references}

\end{document}